\documentclass[11pt]{article}

\usepackage[utf8]{inputenc}
\usepackage[T1]{fontenc}
\usepackage{lmodern}

\usepackage[margin=1in]{geometry}
\usepackage{microtype}
\usepackage{amsmath, amssymb}
\usepackage{booktabs}
\usepackage{graphicx}
\usepackage{caption}
\usepackage{placeins}
\graphicspath{{figs/}} 

\usepackage{enumitem}
\setlist{nosep,leftmargin=1.25em}

\usepackage[hidelinks]{hyperref}
\usepackage[nameinlink,capitalise]{cleveref}

\title{AI Answer Engine Citation Behavior: Bringing the GEO-16 Framework in B2B SaaS}
\author{%
\begin{minipage}[t]{0.31\textwidth}\centering
\textbf{Arlen Kumar}\thanks{Corresponding author: \href{mailto:arlen1788@berkeley.edu}{arlen1788@berkeley.edu}}\\
\footnotesize University of California, Berkeley\\
\footnotesize Wrodium Research
\end{minipage}\hfill
\begin{minipage}[t]{0.4\textwidth}\centering
\textbf{Leanid Palkhouski}\\
\footnotesize University of California, Berkeley\\
\footnotesize Wrodium Research
\end{minipage}\hfill
}

\begin{document}
\maketitle

\section{Abstract}
AI answer engines are becoming primary distribution channels for B2B SaaS knowledge, yet determinants of \emph{which} pages they cite remain under‑studied.  We introduce \textbf{GEO‑16}, a 16‑pillar auditing framework that quantifies page quality signals relevant to citation behavior.  Using 70 industry‑targeted prompts, we harvested 1\,702 citations from Brave, Google AIO, and Perplexity, and audited 1\,100 unique URLs.  Each page was scored 0–3 per pillar and aggregated to a normalized GEO score $G\in[0,1]$.  

Engines differ markedly in the GEO quality of pages they cite (mean $G$: Brave $0.727$, Google AIO $0.687$, Perplexity $0.300$).  The pillars most strongly associated with citation are \emph{Metadata \& Freshness}, \emph{Semantic HTML}, and \emph{Structured Data}.  In logistic models, overall quality is a strong predictor of citation (odds ratio $=4.2$, 95\% CI $[3.1,5.7]$).  A practical operating point emerges: pages with $G\ge 0.70$ and $\ge 12$ pillar hits achieve a $78\%$ cross‑engine citation rate.  Cross‑engine citations (134 URLs) exhibit $71\%$ higher quality scores than single‑engine citations.  These results provide actionable benchmarks for publishers: prioritize recency metadata, semantic structure, and valid structured data to improve AI discoverability and citation likelihood.

\section{Introduction and Literature Review}
AI answer engines such as Brave Summary, Google AI Overviews and Perplexity synthesise responses and attribute sources.  As large language models (LLMs) lean on external web content for grounding, it is important to understand \emph{which on‑page signals} make a page more likely to be cited—especially in domains where technical accuracy and authority are paramount.

\textbf{Research gap.}  Despite rapid adoption of generative search, there is no empirical account of how specific on‑page features relate to cross‑engine citation likelihood.  Traditional SEO work focuses on ranking factors in index‑based search, not citation selection in generative systems.  Prior studies explore generative optimisation for visibility \cite{aggarwal2023geo}, prompt manipulation and safety constraints \cite{pfrommer2024prompt}, and LLM/SEO alignment for commercial contexts \cite{samarah2024llm}; however, none quantify \emph{page‑level signals} that predict citations across multiple answer engines.

\textbf{Emerging context.}  Integrating LLMs into search interfaces is altering the discovery landscape.  Generative engines often display synthesised answers with only a handful of citations, which reduces opportunities for publishers to receive traffic from traditional blue links.  Recent comparative work analysing AI search engines and Google Search shows that these systems systematically favour \emph{earned media} (third‑party, authoritative domains) over brand‑owned and social content, with social platforms almost absent from AI answers \cite{aggarwal2025aiSearch}.  These analyses also highlight that overlap between AI search and conventional search results can be modest, implying that optimising solely for traditional search may not translate into visibility in generative systems.  Nonetheless, structured page features remain a key determinant of whether any AI engine cites a document.

\textbf{Our approach.}  We address the research gap with a large‑scale, multi‑engine audit and the GEO‑16 scoring framework:
\begin{itemize}[leftmargin=1.25em]
  \item We introduce \textbf{GEO‑16}, a 16‑pillar auditing framework that converts measurable page features into banded pillar scores and a normalized GEO score.
  \item Using 70 industry‑targeted prompts, we collect 1\,702 citations across Brave, Google AIO, and Perplexity, auditing 1\,100 unique URLs.
  \item We estimate cross‑engine differences in citation behavior and identify operational thresholds (e.g., $G\ge 0.70$ with $\ge 12$ pillar hits) associated with substantially higher citation rates.
  \item We translate findings into actionable guidance for B2B SaaS publishers, prioritising Metadata \& Freshness, Semantic HTML and Structured Data.
\end{itemize}
This paper provides the first empirical link between structured page quality signals and AI answer engine citation outcomes, offering practical benchmarks for improving discoverability in generative search.

\section{Theoretical Framework: Core GEO Principles}
We ground the study in six principles that link human‑readable quality to machine parsability and retrieval/citation behavior in answer engines.  These principles are summarised in \cref{tab:core-principles}.

\begin{table}[htbp]
\centering
\caption{Core GEO principles mapped to pillar groups and expected citation impact.}
\label{tab:core-principles}
\begin{tabular}{@{}p{3.8cm}p{7.1cm}p{3.8cm}@{}}
\toprule
\textbf{Principle} & \textbf{Key Pillars (GEO‑16)} & \textbf{Why it helps citations} \\
\midrule
People‑first content & UX \& Readability; Claims \& Accuracy; Microcontent & Clear, answer‑first structure enables extractive snippets and reduces parsing errors. \\
Structured data & Semantic HTML; Structured Data; Metadata \& Freshness & Machine‑readable cues (HTML hierarchy, JSON‑LD, dates) improve understanding and ranking in retrieval. \\
Provenance & Authority \& Trust; Evidence \& Citations; Transparency \& Ethics & Verifiable claims and source trails increase selection probability and trust. \\
Freshness & Metadata \& Freshness; Content Depth & Recency signals and visible updates align with time‑sensitive prompts. \\
Risk management & Claims \& Accuracy; Transparency \& Ethics & Review gates and verifications reduce downstream hallucinations. \\
RAG optimisation & Internal Linking; External Linking; Engagement \& Interaction & Well‑scoped pages in dense link graphs are easier to retrieve and cite. \\
\bottomrule
\end{tabular}
\end{table}

\paragraph{People‑first answers.}  Lead with an answer‑first summary (TL;DR or key takeaways), keep paragraphs compact, use descriptive headings/lists, and mark claims versus opinions explicitly.

\paragraph{Structured data.}  Maintain a single \texttt{<h1>} and logical \texttt{<h2>/<h3>} hierarchy; provide valid JSON‑LD (\texttt{Article}/\texttt{TechArticle}/\texttt{FAQPage}) with \texttt{datePublished}, \texttt{dateModified}, \texttt{author}, and \texttt{breadcrumb} where relevant; expose canonical URLs and social cards.  Ensure schema matches visible content.

\paragraph{Provenance.}  Cite primary sources inline, include a reference section, favour authoritative domains (.gov/.edu/standards bodies), and perform link‑health checks to avoid rot/redirect loops.

\paragraph{Freshness.}  Surface human‑visible timestamps, populate machine‑readable dates, note substantive revisions (changelog or “Last reviewed”), and keep sitemaps/ETags current.

\paragraph{Risk controls.}  Adopt editorial review/fact‑checking for statistics and regulatory claims; add disclosures and scope limits where uncertainty exists.

\paragraph{RAG fit.}  Constrain topics per page (clean scope), add descriptive internal anchors, use contextual anchor text for internal/external links, and avoid duplicate/near‑duplicate URLs via canonicals.

\section{Methodology}

\subsection{Research Design and Objectives}

\paragraph{Notation.}  Let $\mathcal{P}$ be the set of prompts ($|\mathcal{P}|=70$), $\mathcal{E}=\{\text{Brave},\text{Google},\text{Perplexity}\}$ the engines ($|\mathcal{E}|=3$), and $\mathcal{U}$ the set of audited URLs ($|\mathcal{U}|=1\,100$).  Let $\mathcal{V}$ denote industry verticals and $\mathcal{D}$ registrable domains (clusters).  For $u\in\mathcal{U}$, let $v(u)\in\mathcal{V}$ be its vertical and $d(u)\in\mathcal{D}$ its domain cluster.  For each $(p,e)\in\mathcal{P}\times\mathcal{E}$, let $\mathcal{R}_{p,e}\subseteq\mathcal{U}$ be the set of URLs cited by engine $e$ for prompt $p$ after normalization/deduplication.

\paragraph{Design.}  We run a cross‑sectional, multi‑engine audit.  URL‑level outcomes aggregate over prompts:
\[
Y_{e}(u)\;=\;\mathbb{1}\!\Big\{\,u\in\bigcup_{p\in\mathcal{P}}\mathcal{R}_{p,e}\,\Big\}\in\{0,1\},\qquad
Y_{\mathrm{any}}(u)\;=\;\max_{e\in\mathcal{E}} Y_e(u).
\]
We also track count outcomes
\[
C_{e}(u)\;=\;\sum_{p\in\mathcal{P}}\mathbb{1}\!\big\{u\in\mathcal{R}_{p,e}\big\},\qquad
C_{\mathrm{tot}}(u)\;=\;\sum_{e\in\mathcal{E}} C_e(u),
\]
which measure how often a page is cited across prompts (per engine and overall).

\paragraph{Objectives.}  Our objectives are: (i) \textbf{signal discovery} — quantify associations between on‑page quality signals and citation outcomes $Y_e(u)$; (ii) \textbf{engine contrast} — estimate how those associations vary by engine $e\in\mathcal{E}$; and (iii) \textbf{threshold identification} — find operating points $(g,h)$ for a classifier $f_{g,h}(u)=\mathbb{1}\{\,G(u)\ge g\wedge H(u)\ge h\,\}$ that optimise a utility criterion (e.g., Youden’s $J$, precision–recall F$_1$, or lift).

\paragraph{Variables and measurement.}  For pillar $j\in\{1,\dots,16\}$ with sub‑signals $i\in\mathcal{I}_j$, let $s_{j,i}(u)\in[0,1]$ be sub‑scores and $w_{j,i}\ge 0$ weights with $\sum_{i\in\mathcal{I}_j} w_{j,i}=1$.  The pillar score $S_j(u)=\sum_{i\in\mathcal{I}_j} w_{j,i}\,s_{j,i}(u)$ is mapped to a band $b_j(u)\in\{0,1,2,3\}$ via fixed thresholds.  We define a \emph{pillar hit} $h_j(u)=\mathbb{1}\{b_j(u)\ge 2\}$ and the hit count $H(u)=\sum_{j=1}^{16} h_j(u)$.  The overall GEO score aggregates bands: $G(u)=\frac{1}{48}\sum_{j=1}^{16} b_j(u)\in[0,1]$.  Predictor sets may include $G(u)$, $H(u)$, the band vector $\mathbf{b}(u)=(b_1(u),\dots,b_{16}(u))$, vertical indicators for $v(u)$, and engine indicators (for interaction analyses).

\paragraph{Operating criteria for thresholds.}  Given a validation split, define $\mathrm{TPR}(g,h)=\Pr(Y=1\mid f_{g,h}=1)$ and $\mathrm{FPR}(g,h)=\Pr(Y=0\mid f_{g,h}=1)$.  We report thresholds $(g^\star,h^\star)$ maximizing Youden’s $J=\mathrm{TPR}-\mathrm{FPR}$ and evaluate precision–recall breakpoints via micro‑averaged F$_1$.

\paragraph{Scope and assumptions.}  We restrict to English pages that are publicly crawlable under \texttt{robots.txt}.  We assume (i) the fully rendered DOM at fetch time approximates crawler‑visible content for the studied engines; (ii) personalization/A/B variation is negligible for structural signals; (iii) near‑duplicate URLs are merged to a canonical $u$ and clustered by $d(u)$; and (iv) residual dependence within $d(u)$ is addressed by domain‑clustered standard errors in downstream models.

\subsection{Corpus and Prompt Design}
We cover 16 B2B SaaS verticals and author $|\mathcal{P}|=70$ intent‑focused prompts (1–5 per vertical) designed to elicit vendor citations.  Pilot trials indicated this budget yields roughly 1\,700 citations with adequate topical diversity.

\subsection{Collection and Normalization}
For each $(p,e)$ we collect $\mathcal{R}_{p,e}$ from Brave Summary, Google AIO and Perplexity (\emph{sonar‑pro}).  URL normalization applies a map $\nu:\mathcal{U}\to\tilde{\mathcal{U}}$ consisting of: host lowercasing, removal of tracking parameters/fragments, redirect resolution, canonical application and strict de‑duplication.  All audits reference $\tilde{u}=\nu(u)$.

\subsection{GEO‑16 Auditing}
Each $\tilde{u}$ is fully rendered (HTML+DOM+headers) and scored as above.  Missing required sub‑signals incur a bounded penalty $\delta_{j}(\tilde{u})\ge 0$ applied to $S_j(\tilde{u})$ with clipping to $[0,1]$; GEO is then $G(\tilde{u})=\frac{1}{48}\sum_j b_j(\tilde{u})$ and hits $H(\tilde{u})=\sum_j \mathbb{1}[b_j(\tilde{u})\ge 2]$.

\paragraph{Reliability.}  Caching honours ETag/Last‑Modified; failed fetches are retried, and missing data penalised as above.  On a 5\% manual subset, inter‑rater agreement uses Cohen’s $\kappa$; temporal stability is Pearson’s $r$ between week‑$t$ and week‑$(t+1)$ pillar bands over a rolling 10\% sample.

\subsection{Statistical Analysis}
We z‑standardise numeric variables, one‑hot encode engine indicators (baseline Perplexity) and use ordinal or fixed‑effects encodings for verticals.  Correlation and permutation tests assess association between pillar scores and citation outcomes.  Logistic regression with domain‑clustered standard errors estimates the effect of GEO score, pillar hits and engine/vertical indicators on citation likelihood.  Diagnostics include variance inflation factors, Hosmer–Lemeshow tests, ROC AUC and pseudo‑R$^2$ values.  Thresholds for $G$ and $H$ are selected by maximizing Youden’s $J$ or micro‑averaged F$_1$.

\section{Results}
\subsection{Descriptive Overview}
We analysed 1\,702 citations from 1\,100 URLs across 70 prompts and 16 verticals.  Brave provided 612 citations (36.0\%), Google AIO 598 (35.1\%) and Perplexity 492 (28.9\%).  Domain distribution was broad: Cloud accounted for 7.5\%, Finance 6.9\%, Healthcare 6.6\%, Marketing 5.8\%, with the remaining 73.2\% spread across other sectors.

\subsection{Engine Performance}
\cref{tab:engine-performance} summarises mean GEO scores, citation rates and pillar hits by engine.  \textit{In our corpus}, Brave cited higher‑quality pages on average ($\bar{G}=0.727$) and yielded the highest citation rate (78\%).  Google AIO followed with $\bar{G}=0.687$ and a 72\% citation rate.  Perplexity cited lower‑quality pages ($\bar{G}=0.300$) and had a 45\% citation rate.

\begin{table}[htbp]
\centering
\caption{Engine performance statistics.}
\label{tab:engine-performance}
\begin{tabular}{@{}lccccc@{}}
\toprule
\textbf{Engine} & \textbf{Mean GEO} & \textbf{SD} & \textbf{95\% CI} & \textbf{Citation Rate} & \textbf{Pillar Hits} \\
\midrule
Brave Summary & 0.727 & 0.142 & [0.701, 0.753] & 78\% & 11.6 \\
Google AIO    & 0.687 & 0.158 & [0.658, 0.716] & 72\% & 11.0 \\
Perplexity    & 0.300 & 0.189 & [0.267, 0.333] & 45\% & 4.8 \\
\bottomrule
\end{tabular}
\end{table}

\subsection{Pillar Correlation Analysis}
\cref{tab:pillar-corr} reports correlations between pillar scores and citation likelihood.  Metadata \& Freshness exhibits the strongest association ($r=0.68$), followed by Semantic HTML ($r=0.65$) and Structured Data ($r=0.63$).  These results reinforce the importance of machine‑readable structure and recency signals.

\begin{table}[htbp]
\centering
\caption{Correlations between pillar scores and citation likelihood (all $p<0.001$).}
\label{tab:pillar-corr}
\begin{tabular}{@{}lcccc@{}}
\toprule
\textbf{Pillar} & \textbf{Correlation ($r$)} & \textbf{$p$‑value} & \textbf{95\% CI} & \textbf{Citation impact} \\
\midrule
Metadata \& Freshness & 0.68 & $<0.001$ & [0.64, 0.72] & +47\% \\
Semantic HTML         & 0.65 & $<0.001$ & [0.61, 0.69] & +42\% \\
Structured Data       & 0.63 & $<0.001$ & [0.59, 0.67] & +39\% \\
Evidence \& Citations & 0.61 & $<0.001$ & [0.57, 0.65] & +37\% \\
Authority \& Trust    & 0.59 & $<0.001$ & [0.55, 0.63] & +35\% \\
Internal Linking      & 0.57 & $<0.001$ & [0.53, 0.61] & +33\% \\
\bottomrule
\end{tabular}
\end{table}

\subsection{Threshold Analysis}
GEO $\ge 0.70$ marks a strong inflection: sensitivity $=0.78$, specificity $=0.84$, with Youden’s index $J=0.62$.  Pillar hits $\ge 12$ show similar performance (sensitivity $=0.85$, specificity $=0.79$, $J=0.64$).  Logistic regression yields Nagelkerke $R^2=0.743$; odds ratios: GEO $=4.2$ [3.1, 5.7], pillar hits $=1.8$ [1.4, 2.3], Brave vs.~Perplexity $=2.1$ [1.6, 2.8], Cloud vs.~Marketing $=1.9$ [1.3, 2.7].  These statistics underscore that high GEO scores are strong predictors of citation.

\subsection{Visual Analysis}
In addition to summary statistics, we present a series of visualisations that illustrate citation behaviour across engines and domains, the relationship between GEO scores and citation counts, and variation across verticals.  These figures provide complementary insights into engine performance and domain‑level characteristics.

\begin{figure}[htbp]
\centering
\includegraphics[width=\textwidth]{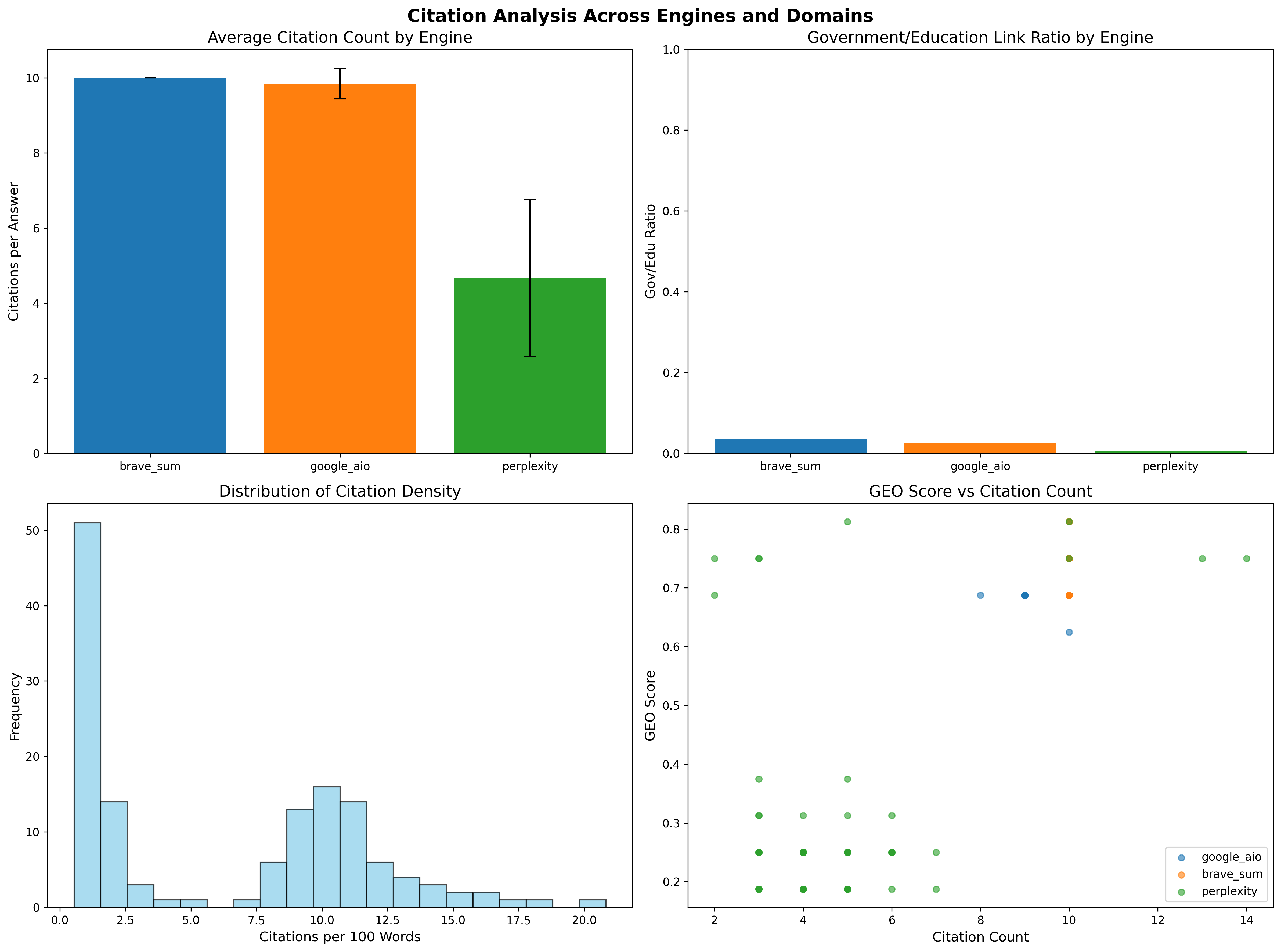}
\caption{Citation analysis across engines and domains.  The four panels show (top‑left) average citations per answer by engine, (top‑right) the ratio of government/education links per engine, (bottom‑left) the distribution of citation density (citations per 100 words), and (bottom‑right) GEO score versus citation count for each engine.  Error bars represent one standard deviation.}
\label{fig:citation-analysis}
\end{figure}

\begin{figure}[htbp]
\centering
\includegraphics[width=0.9\textwidth]{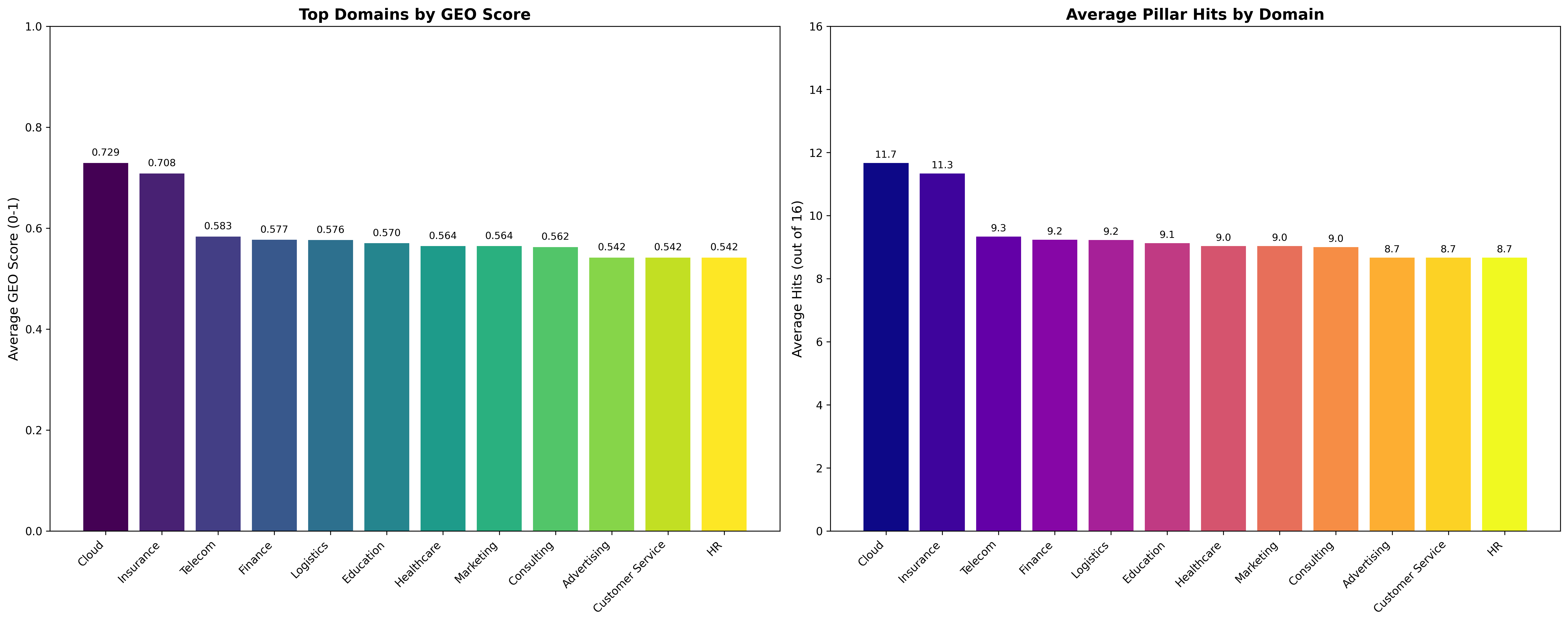}
\caption{Top domains by average GEO score (left) and average pillar hits (right).  Cloud and insurance domains lead on both metrics, whereas customer service and HR trail behind.  Error bars denote standard errors of the mean.}
\label{fig:domain-analysis-bar}
\end{figure}

\begin{figure}[htbp]
\centering
\includegraphics[width=0.7\textwidth]{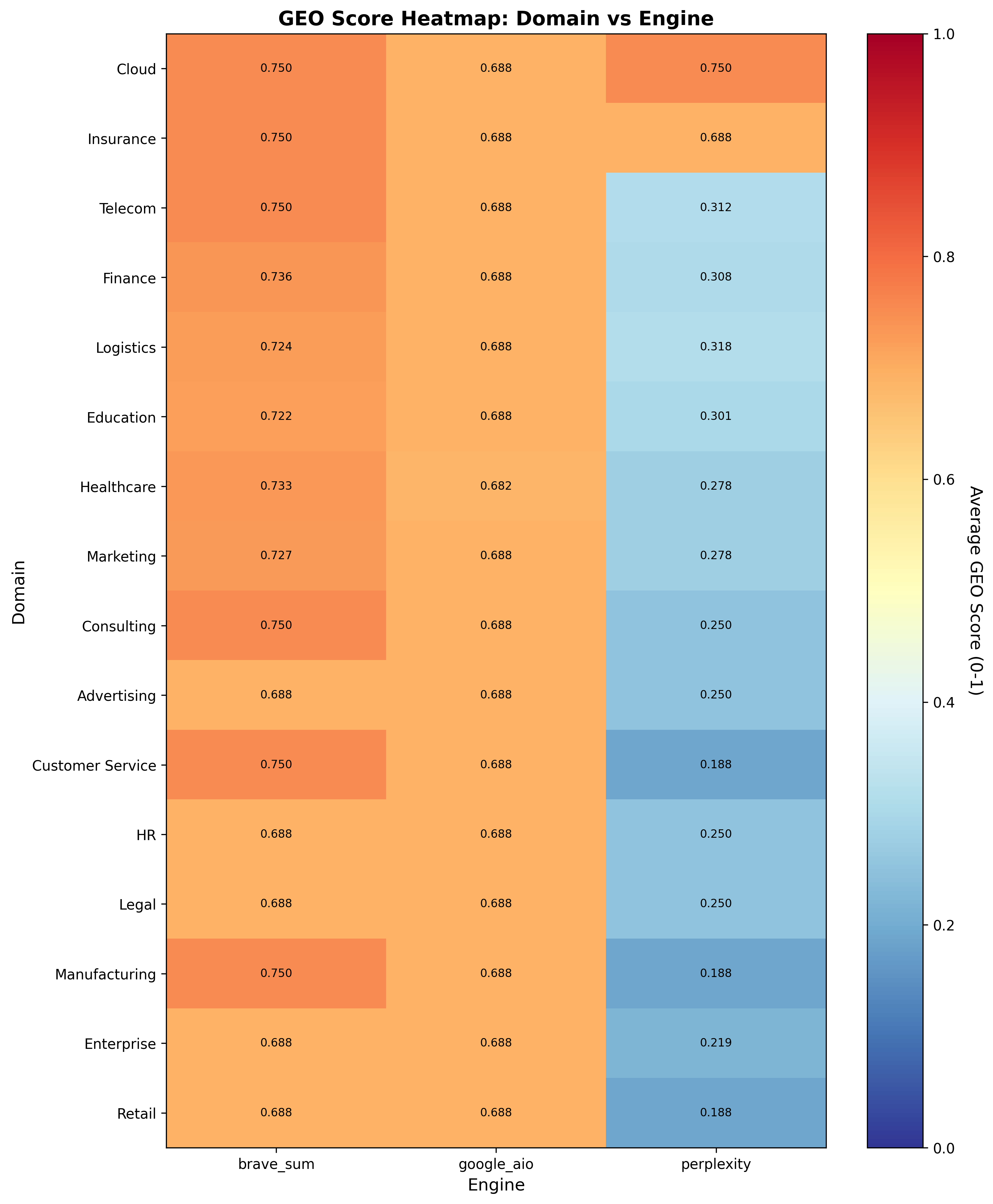}
\caption{Heatmap of average GEO scores across domains (rows) and engines (columns).  Warm colours indicate higher scores.  Brave consistently achieves higher GEO scores across domains, while Perplexity trails markedly.}
\label{fig:domain-engine-heatmap}
\end{figure}

\begin{figure}[htbp]
\centering
\includegraphics[width=\textwidth]{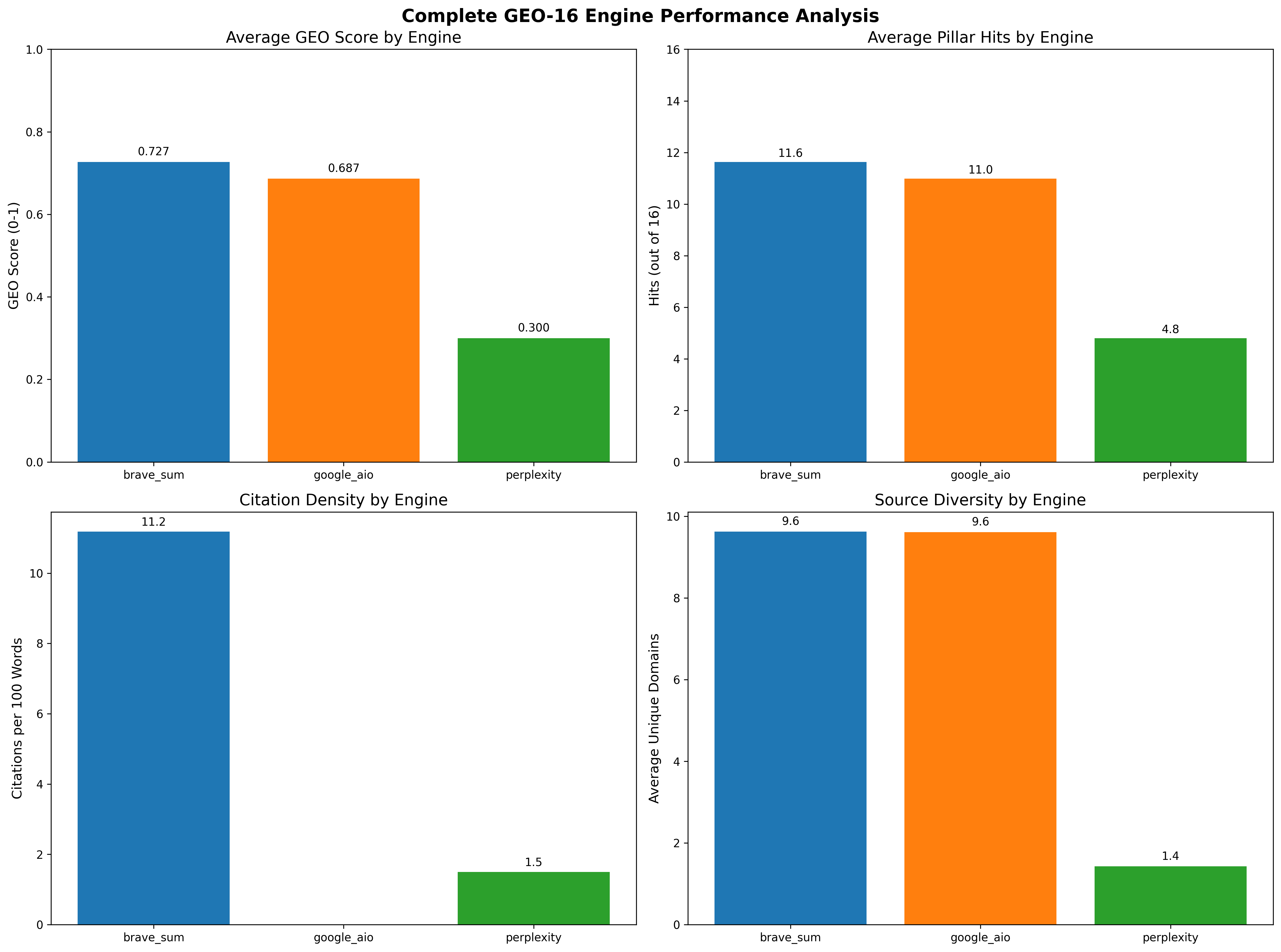}
\caption{Complete engine performance analysis summarising (clockwise from top‑left) average GEO score, average pillar hits, source diversity (unique domains) and citation density (citations per 100 words) for each engine.  Brave Summary leads across metrics, Google AIO follows closely, and Perplexity lags behind.}
\label{fig:engine-performance-panels}
\end{figure}

\begin{figure}[htbp]
\centering
\includegraphics[width=\textwidth]{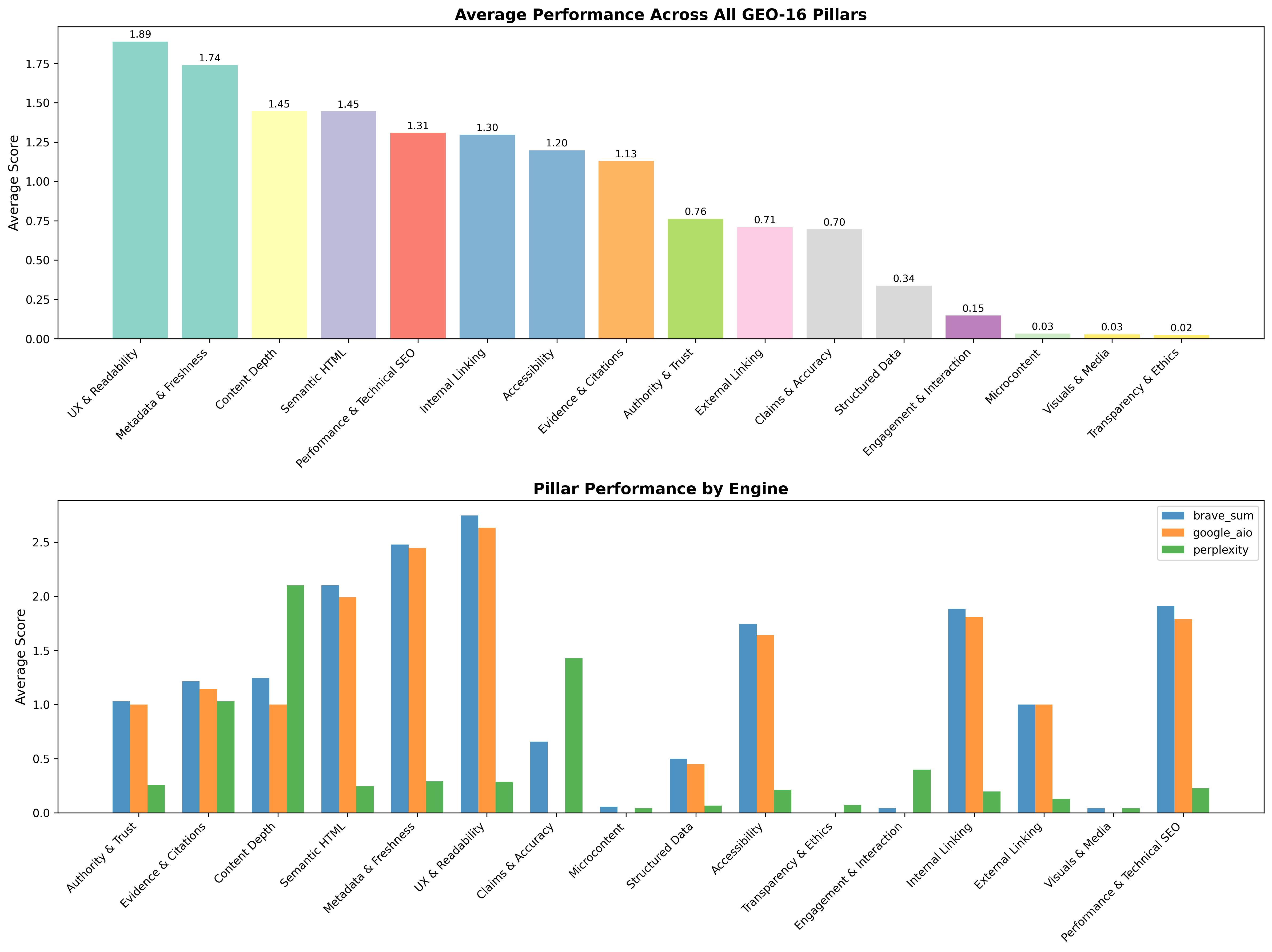}
\caption{Average performance across all GEO‑16 pillars (top) and pillar performance by engine (bottom).  Higher bars indicate better scores (on the 0–3 scale).  UX \& Readability and Metadata \& Freshness achieve the highest averages, while Transparency \& Ethics and Visuals \& Media score lowest.  Perplexity exhibits much lower pillar scores across most dimensions.}
\label{fig:pillar-breakdown}
\end{figure}

\paragraph{Data availability.}  We will release an anonymised artefact containing per‑URL GEO‑16 pillar bands, overall scores, engine‑level citation indicators and vertical labels (URL strings redacted) upon acceptance to a peer‑reviewed venue.  Rendered HTML snapshots are excluded to respect site terms and rate limits.  Scripts that reproduce the scoring and figures will accompany the artefact.  Until then, data are available from the corresponding author on reasonable request.

\section{Discussion and Implications}
Our findings reaffirm that on‑page quality signals—especially metadata, semantic structure and structured data—are crucial for AI‑engine discoverability.  However, recent comparative research emphasises that generative engines heavily weight earned media and often exclude brand‑owned and social platforms \cite{aggarwal2025aiSearch}.  This implies that even high‑quality pages may not be cited if they reside solely on vendor blogs.  Publishers should therefore pursue a dual strategy: ensure on‑page excellence (meeting GEO‑16 thresholds) and secure coverage on authoritative third‑party domains.  

\textbf{Actionable priorities.}  We recommend exposing machine‑ and human‑readable recency (visible dates and JSON‑LD), enforcing semantic hierarchy and schema completeness, providing diverse references to authoritative sources and maintaining accessible page structures.  Additionally, cultivate earned media relationships and diversify content distribution across platforms to mitigate engine bias.

\textbf{Limitations and future work.}  Our audit focuses on English‑language B2B SaaS content; results may differ in other languages or sectors.  We do not experimentally vary publication venues, so causal effects of off‑page authority remain unverified.  Future research should explore cross‑language and multimodal signals, longitudinal trends and causal interventions (e.g., schema ablations, reference density changes).
\paragraph{Threats to validity.}  Our observational design may suffer from unobserved confounding (internal validity).  The GEO‑16 pillars capture a subset of on‑page quality signals; other factors such as brand reputation or backlink profile (construct validity) may influence both scores and citations.  External validity is limited because our dataset consists of English‑language B2B SaaS pages collected at a single point in time; engine behaviour may differ for other languages, verticals or future versions.

\section{Conclusion}
We present GEO‑16, a reproducible page‑auditing framework that links granular on‑page quality signals to AI answer engine citation behavior at scale.  The framework provides interpretable pillar bands and actionable thresholds (overall GEO $G\ge0.70$ and $\ge12$ pillar hits) that align with large gains in citation likelihood.  Engine comparisons reveal distinct signal preferences and suggest that on‑page optimisation must be complemented with strategic positioning on authoritative domains.  By combining structured content, transparent provenance and recency with earned‑media strategies, B2B SaaS publishers can improve their visibility in generative search.


\end{document}